# Explainable AI for Data-Driven Design of High-Dimensional Predictive Studies

**Junyu Yan[*1,2,3], Damian Machlanski[*1,2], Kurt Butler[1,2], Panagiotis Dimitrakopoulos[1,2], Ewen M Harrison[4], Bruce Guthrie[3], Sotirios A Tsaftaris[1,2]**

1. School of Engineering, University of Edinburgh, Edinburgh, UK
2. Causality in Healthcare AI Hub (CHAI), UK
3. Advanced Care Research Centre, Usher School of Population Health Sciences, University of Edinburgh, Edinburgh, UK
4. Centre for Medical Informatics, Usher School of Population Health Sciences, University of Edinburgh, Edinburgh, UK

Corresponding Author: Damian Machlanski (d.machlanski@ed.ac.uk)

## Abstract

Predictive modelling is important for health data analysis and data-driven clinical decision-making. However, predictive studies are challenging to design optimally by hand when tens or even hundreds of features require selection, transformation, or interaction modelling. While complex machine learning models offer high performance, their "black-box" nature limits the clinical trust, transparency, and interpretability required for decision-making. We developed and evaluated an Exploratory AI Recommender that provides data-driven recommendations to improve predictive performance of existing interpretable statistical models. The developed framework uses flexible AI modelling to capture complex data patterns and explainable AI techniques to translate the patterns into three recommendation types: feature exclusion, non-linear terms, and feature interactions. We evaluated the framework by comparing predictive performance of a baseline (i.e., no interactions or non-linear terms) Cox Proportional Hazards (CPH) model against an augmented CPH incorporating recommendations suggested by our method. The primary analysis predicts the time to the first occurrence of a fall or related injury in 245,614 patients. Our method recommended excluding 23 features, including non-linear terms for two features, and including 221 suggested feature interactions. The C-index improved from 0.805 (95% CI 0.798-0.812) to 0.815 (95% CI 0.809-0.822), and so did calibration (intercept: -0.006 to 0.003; slope: 1.063 to 0.950). All recommendations were supported by existing literature. The method also proved effective on two additional public datasets, demonstrating wider applicability. The proposed Exploratory AI Recommender demonstrates the potential of explainable AI and data-driven study design to improve the process of developing, and the performance of high-dimensional transparent predictive models.

[*] The two authors contributed equally.

# Introduction

Predictive modelling is important in healthcare as it enables proactive clinical decisions by forecasting future health outcomes from patient data. A prime example is models for predicting 10-year cardiovascular risk including PREVENT[1] and QRISK,[2] which support clinician and patient decision-making about initiation of preventive medications. Historically, clinical predictive modelling has been the domain of standard statistical methods. Techniques such as Cox Proportional Hazards (CPH) have served medical research for decades.[3,4] They are computationally efficient, mathematically rigorous, and, crucially, intrinsically interpretable.

However, as the dimensionality (i.e., number of variables) of healthcare data increased (e.g., electronic healthcare records), the limitations of these classical approaches have become increasingly apparent. Standard statistical methods, while trustworthy, often rely on linearity assumptions that may not reflect biological reality. In clinical practice, the risk factor-outcome relationship is rarely a straight line. For instance, the risk associated with body mass index or blood pressure may follow a U-shaped curve (i.e., non-linearity), or the impact of a specific biomarker may be additionally dependent on a patient's age or genetic profile (i.e., interaction). Capturing these nuances through "feature engineering", the manual process of transforming variables and specifying interactions, becomes intractable as data grows. For example, with only 100 features, the number of two-way interactions reaches 4 950 possibilities, making it nearly impossible for researchers to manually identify every clinically relevant synergy.

Modern Artificial Intelligence (AI) predictors have demonstrated the capacity to outperform classical approaches in a variety of clinical tasks.[5] These models excel at "end-to-end learning," where the model learns expressive feature representations directly from raw data without the need for manual feature engineering.[6,7] This allows AI models to capture non-linearities and high-order interactions that would be invisible to standard techniques. However, this performance gain comes at a significant cost: the loss of clinical "sense-checking". When a model functions as a "black-box", clinicians cannot verify if a prediction is based on sound pathophysiology or mere statistical noise and "shortcuts" in the dataset.[8] While some argue that high predictive performance should take precedence over interpretability in clinical AI,[9,10] this lack of transparency prevents clinicians from explaining *why* a patient is at risk, hindering the ability to identify high-risk profiles that warrant further clinical investigation and undermining the trust required for clinical adoption in high-stakes environments.[11–13]

To address the black-box problem, the field of explainable AI has emerged.[14] One line of explainable AI research is on attribution, the process of assigning "credit" to input features for a model's prediction using methods like SHAP (SHapley Additive exPlanations).[15] Explainable AI methods have been successfully used to analyze risk predictors,[16–19] to identify biomarkers in diseases,[20,21] or factors in cardiovascular risk.[22,23]

This study explores the potential of explainable AI not merely as a tool for post-hoc justification of black-box predictions, but as a mechanism to design better white-box models, a process we refer to as data-driven study design. While previous work investigated automated feature engineering with simple models,[24] and augmented Cox models with explainable AI-derived new predictors,[25] it remains unclear whether explainable AI can automate all three feature engineering tasks (feature selection, non-linear terms, interaction terms) to improve the performance of conventional clinical models on high-dimensional health data. Our method aims to bridge this gap by gaining predictive strength of complex AI models while retaining the simplicity of conventional predictors. By using explainable AI to "discover" the complex relationships and then embedding them back into transparent statistical tools, we provide clinicians with a model that is as powerful as modern AI but as auditable as a standard risk score.

# Methods

Our study proposes an Explainable AI-based Exploratory AI Recommender (*Recommender* henceforth) designed to systematically uncover inherent data relationships and optimize the design of predictive cohort studies. The methodology follows three stages (Figure 1): (1) establishing a baseline predictive model, representing the initial analytic stage of a research team; (2) deploying our Recommender to generate actionable insights and recommendations for refining study design; and (3) evaluating the Recommender's utility by quantifying performance gains of models augmented with these recommendations. We apply this methodology to a clinical case study: predicting the risk of falls or related injuries (FRI) using DataLoch data.

## Data Source and Study Population

The analysis was conducted on a high-dimensional clinical dataset provided by Lothian DataLoch, a secure, regional data repository linking primary and secondary care data in South East Scotland.[26] This study included all individuals aged over 50 years, permanently registered for at least one year with a Lothian General Medical Practice (GP) that contributes to the DataLoch dataset in the period 01/01/2015-28/02/2020. This pre-COVID-19 period was specifically chosen to mitigate potential biases on FRI that might arise from pandemic-related changes in healthcare-seeking behavior. Individuals were excluded if data for age, sex, or the index of multiple deprivation were missing. Patients left the cohort at the earliest of death, GP de-registration, an outcome event, or the study end date (February 28, 2020). The dataset was randomly split into 80% training (~196 491 rows) and 20% held-out test data (~49 123 rows). This study follows the Transparent Reporting of a Multivariable Prediction Model for Individual Prognosis or Diagnosis (TRIPOD+AI) reporting guidelines.[27]

## Outcomes

The primary outcome was the time from cohort entry to the first occurrence of FRI, where related injuries were defined as low energy fractures (hip, wrist, and proximal humerus). Diagnosis of FRI was defined by clinical codes from primary care (Read version 2), hospital admissions (ICD-10), and emergency department records (Supplementary Table 1).

## Design, Implementation and Evaluation of the Exploratory AI Recommender

First, we established a baseline predictive model using the standard multivariable Cox Proportional Hazards model. This model represents the starting point of a conventional, literature-driven predictive modeling study. Following the knowledge-driven approach, we selected 58 predictors from approximately 400 available features based on established literature. These include four patient demographic characteristics, three lifestyle risk factors, 27 comorbidities, electronic Frailty Index category,[28] and exposure to 23 Fall-Risk-Increasing Drugs (list of features in Supplementary Table 4; see Supplementary Note 1 for further details). The baseline model was fitted on the training data using all 58 candidate predictors, deliberately omitting advanced techniques such as feature selection, non-linear terms, or interaction terms. Its performance on the held-out test set served as the baseline for evaluating Recommender-driven refinements.

Second, we deployed the Recommender to systematically explore underlying data relationships and generate concrete recommendations for refining the above knowledge-driven predictive study design. This began with fitting an advanced exploratory AI model to the training data to capture complex, non-linear relationships and potential interactions between all features and outcomes. For this purpose, we employed Random Survival Forest[29] (RSF) as the exploratory AI model. Unlike typical machine learning applications, the RSF was used purely to capture high-dimensional data structures rather than for final clinical prediction. Next, the RSF model was interpreted using explainable AI techniques, such as SHAP,[15] to translate "black-box" patterns into actionable insights. SHAP quantifies each feature's contribution (termed *feature attributions* (FAs)) to the RSF's prediction for each patient (see Supplementary Note 2). Standard FA analysis often relies on population level averaging, which can obscure risk factors that are clinically significant only within specific subgroups. To resolve this, we performed an extreme-group FA analysis that targets two distinct risk cohorts: a low-risk and high-risk cohort. The low-risk cohort included individuals with RSF predicted risk within a predefined margin (Euclidean distance < 0.05 times SD of the model's predictions) of the mean score for patients with no FRI events. The high-risk cohort was similarly defined relative to the mean score for patients with a fall event. The FAs from these cohorts were then processed to derive modeling recommendations. While computing FAs might be more intensive on average than training a baseline Cox model, this exploratory phase is a one-time cost that precedes the deployment of the more efficient augmented model.

Our Recommender offers the following three types of recommendations (Figure 2): (1) **Feature exclusion (Figure 2; E).** A feature was recommended for exclusion if the upper bound of the 95% CI of its mean absolute FA did not exceed a data-driven significance threshold. The threshold was defined as 5% of the standard deviation of mean absolute FA values across all candidate features, a heuristic to distinguish features with negligible contributions; (2) **Non-linear relationships (Figure 2; D).** We calculated the Pearson correlation coefficient $r$ between the feature values and their corresponding FAs for each feature. For any feature with a weak correlation ($|r| < 0.1$), we recommended modelling its effect using non-linear terms (e.g., quadratic terms, splines); (3) **Feature interactions (Figure 2; A-C).** We employed an iterative stratification analysis on FAs to identify potential feature interactions, guided by three types of data patterns (see Figure 2). Once stratified, FAs were produced for each stratum (i.e., Step 2 above repeated for each stratum). A Wilcoxon rank-sum test was then employed to compare FA distributions between created strata. Statistically significant differences indicated a recommendation to include a first-order (i.e., two-way) interaction term between the stratifying feature and the statistically different features.

Third, we evaluated the effectiveness of the Recommender. We constructed augmented Cox models that integrated these recommendations (excluding features, adding non-linear terms and interactions). We then compared the predictive performance of these augmented models against the baseline on the held-out test set. Furthermore, the clinical plausibility of all generated recommendations was validated by reviewing existing medical literature.

## Application to Other Datasets

To establish the broader applicability and robustness of our method, we applied the Recommender on two additional, publicly available time-to-event datasets: *GBSG2* (breast cancer) and *ACT* (human immunodeficiency virus).[30,31]

*German Breast Cancer Study Group 2 (GBSG2)*. The data comes from a randomized clinical trial that compared different chemotherapy treatment regimens and their effects on survival in breast cancer patients. This study group consists of 686 samples and 8 features with a 43.6% survival rate. The features include: age, estrogen receptor level (estrec), the number of positive (involved) lymph node (pnodes), progesterone receptor level (progrec), hormonal therapy (horTh), tumour grade (tgrade), menopausal status (menostat), tumor size (tsize).

*AIDS Clinical Trial (ACT)*. This data set contains 1151 HIV-positive patients from a randomized controlled trial that compared different drug regimens. The data set includes 11 features: treatment indictor (tx), treatment group indicator (txgrp), CD4 stratum at screening (strat2), sex, race/ethnicity (raceth), IV drug use history (ivdrug), hemophiliac (hemophil), Karnofsky Performance Scale (karnof), Baseline CD4 count (cd4), Months of prior ZDV use (priorzdv), and age. The endpoint was time to an AIDS-defining event or death, with a censoring rate of 89.4%.

All data sets were accessed via the scikit-survival Python library.[1] For each data set, the cohort was randomly split into a training set (80%) and a test set (20%). The complete Explainable AI-based Exploratory AI Recommender workflow, including fitting a baseline Cox model, training an exploratory (RSF) model with 1000 trees, and applying SHAP to derive recommendations, was performed in the same way as for the primary dataset.

### Statistics and Reproducibility

Baseline characteristics of study samples were described using frequency and percentage, mean and standard deviation (SD). Statistical analyses, model development, and testing were conducted in Python, version 3.12.5. All model performances (i.e., baseline CPH, RSF, and augmented CPH) were quantified by the concordance index (C-index) for discrimination and calibration plots for calibration (intercept and slope). All confidence intervals were 95% percentile intervals generated using 1000 bootstraps. All tests were 2-tailed, and $P < 0.05$ was considered as significant.

### Study Approval and Consent

All data from DataLoch were linked and de-identified by the DataLoch service (Edinburgh, United Kingdom) and analyzed within a Secure Data Environment (reference: DL_2024_030). DataLoch enables access to de-identified extracts of healthcare data from the South-East Scotland region to approved applicants: https://dataloch.org/. The study was reviewed and received approval under delegated authority from a regional National Health Service Research Ethics Committee (REC 22/NS/0093) and Caldicott Guardian. Individual patient consent was not required. Clinical trial number: not applicable.

## Results

The DataLoch cohort included 245 614 patients with a mean (SD) age of 67 (12) years, and included 114 067 (46.4%) females, of whom 891 (0.4%) self-identified as Black, 3444 (1.4%) as Asian, and 207 254 (84.4%) as White. The most prescribed medications were opioid analgesics (188 237 patients [76.7%]), non-steroidal anti-inflammatory drugs (71 078 patients [28.9%]) and angiotensin-converting enzyme inhibitors (66 644 patients [27.1%]), while the most prevalent comorbidities were hypertension (52 330 patients [21.3%]) and cancers (27 371 patients [11.1%]). Supplementary Table 4 contains the patients' characteristics of the study.

### Baseline Performance

The baseline CPH achieved a C-index of 0.805 (95% CI 0.798-0.812) with a calibration slope of 1.063 and intercept of -0.006 on the test set (Figure 3; A). The exploratory RSF model achieved a C-index of 0.820 (95% CI 0.813-0.827) with a calibration slope of 1.142 and intercept of -0.014 (Figure 3; B). The discriminative and well-calibrated performance demonstrated the ability of the exploratory model to capture complex relationships between features.

---

[1] https://scikit-survival.readthedocs.io/en/stable/api/datasets.html

## Generated Recommendations

Figure 4 and Figure 5 detail the process for conducting FA analysis and deriving recommendations. The fall risk scores (derived from the RSF model) and pre-defined margins were used to select low-risk and high-risk cohorts. As a result, 6928 and 2362 patients were selected into the low-risk and high-risk cohort respectively from the initial training set (~196 491 rows). The FA analysis was conducted on these two cohorts separately (Figure 4). This analysis generated several recommendations for model refinement. First, it recommended 22 features for exclusion as their contributions were not significant either in the high-risk or low-risk cohort (Supplementary Figure 1). Second, within the remaining 36 features, it identified two features with a weak correlation between FAs and feature values (low-risk cohort: alcohol status ($|r|$ = -0.03 (P=.29)); high-risk cohort: alcohol status ($|r|$ = -0.10 (P=.11)), age ($|r|$ = 0.01 (P=.86))). These features were recommended as non-linear terms in further modeling. Finally, the analysis identified 221 first-order interactions through stratification on 10 variables (e.g., age, loop diuretics, dementia) following three data stratification patterns (Supplementary Figure 1). Take age as an example, age demonstrated a wide range of FAs; below a certain feature value, it contributed negatively to FRI risks, whereas above that value, it contributed positively. Consequently, age was stratified by feature values to illustrate interactions with other features. Figure 5 shows age-stratified results within the low-risk cohort using a 65-year cutoff (a conventional threshold for older age group and close to the mean age at which FA ≈ 0). 2362 patients were aged under 65, and 4566 were over 65. Several features have significantly different FA distributions between two age strata, particularly in opposing directions (e.g., antipsychotic drugs, stroke). These results suggest that the effects of these factors on FRI risk are modified by age, illustrating their interactions. The identified interactions were recommended as multiplicative terms between the stratified feature and its corresponding interacted features in the further augmented model.

## Quantitative Evaluation of Recommendations

We integrated the three types of recommendations into augmented Cox models independently and cumulatively to evaluate their individual and collective impact (Figure 3). **Independent assessment.** An augmented CPH model incorporating only the feature exclusion recommendations (58 features reduced to 36) demonstrated no degradation in performance compared with the baseline CPH model (ΔC-index, -0.001 (95% CI -0.002-0.000), p=.10). A CPH model incorporating quadratic terms for the three identified non-linear features on top of 58 baseline features achieved a C-index of 0.805 (95% CI, 0.798-0.812) and calibration (slope, 1.063; intercept, -0.005), which slightly improved against the baseline model. An augmented CPH model adding the 221 recommended first-order interactions to the 58 baseline predictors (i.e., no feature exclusion) demonstrated a significant improvement over the baseline model (C-index, 0.811 [95% CI, 0.804-0.818]; ΔC-index, 0.006 [95% CI, 0.004-0.009]; P < 0.001) and better calibration (slope, 0.966; intercept, 0.001). **Cumulative assessment.** The final CPH model, incorporating all 3 types of recommendations (first feature exclusion, then non-linear and interaction terms), achieved a C-index of 0.815 (95% CI, 0.809-0.822) and a calibration slope of 0.950 with an intercept of 0.003. Compared with the baseline CPH model, the final model showed a significant improvement in discrimination (ΔC-index, 0.010; 95% CI, 0.008-0.013; P < 0.001) and calibration.

## Evaluation of Recommendations on Other Datasets

The Recommender generated actionable recommendations that led to significant model performance improvements on two additional datasets (Figure 6 and Figure 7).

**GBSG2**. The baseline Cox model achieved a C-index of 0.665 (95% CI 0.665 – 0.665) on the test set, with a calibration slope of 0.963 and an intercept of 0.106. The corresponding RSF model achieved a C-index of 0.673 (95% CI 0.673-0.673) with the calibration slope of 1.163 and intercept of -0.004 (Figure 6; B).

The feature attribution analysis (Figure 6; A) produced the following recommendations: All 8 features were identified as contributory, and no feature was recommended for exclusion. A non-linear term was recommended for the *age* feature. The framework has identified 14 potential interactions: *age-tsize, age-estrec, estrec-pnodes*, *progrec-age, progrec-pnodes, progrec-estrec, progrec-horTh, progrec-tgrade, progrec-menostat, horTh-menostat, horTh-tsize, horTh-tgrade, horTh-pnodes,* and *horTh-age.*

A refined Cox model incorporating all recommendations achieved a C-index of 0.687 (95% CI 0.687-0.687), which greatly improved the discrimination over the baseline model. The refined model remained well-calibrated (slope: 0.945; intercept: 0.112) (Figure 6; B)

**ACT.** The baseline Cox model achieved the C-index of 0.725 (95% CI 0.725-0.725) on the test set, with a calibration slope of 1.156 and intercept -0.196. The exploratory RSF model achieved C-index of 0.733 (95% CI 0.733-0.733) with a calibration slope of 1.628 and intercept -0.631 (Figure 7; B).

Figure 7; A demonstrated the FA analysis results for the ACT data. Based on the FA analysis, the workflow recommended excluding the feature *hemophil* and suggested including non-linear terms for *karnof* and *age*. It also identified 11 potential interactions: *ivdrug-karnof, cd4-karnof, cd4-raceth*, *priorzdv-ivdrug*, *raceth-ivdrug*, *raceth-priorzdv*, *raceth-sex*, *age-strat2*, *age-priorzdv*, *age-karnof*, *age-raceth*.

After integrating all recommendations, the refined Cox model achieved a C-index of 0.770 (95% CI 0.770-0.770) with the calibration slope of 1.650 and intercept -0.656 (Figure 7; B). The discriminative performance greatly improved over the baseline model, with the calibration performance remaining similar.

These findings demonstrate that our proposed Recommender can consistently generate clinically relevant hypotheses for study design across different data sets and clinical questions.

# Discussion

In this study, we developed and evaluated an Exploratory AI Recommender designed to support predictive study design in high-dimensional clinical datasets. Rather than using AI as an opaque final predication model, the framework uses a flexible explainable-AI module as an exploratory engine to identify data-driven modelling insights that can then be translated into transparent statistical models. In a fall risk case study, the system identified 22 potentially redundant predictors among 58 candidate features, detected two non-linear associations, and uncovered 221 clinically plausible interactions from 4,950 possible feature pairs. An augmented Cox Proportional Hazards model incorporating these findings significantly exceeded the baseline in both discrimination and calibration. By deploying AI strictly as an exploratory mechanism rather than a direct predictor, our framework can help researchers navigate complex feature spaces while preserving the auditability and mathematical transparency required in regulated clinical environments. This explicit separation of automated discovery from final clinical prediction resolves the tension between algorithmic accuracy and interpretability.

The Recommender's outputs align with established literature, confirming frailty and age as dominant FRI predictors[32] and age as non-linearly correlated with risk. It correctly identified known interactions, such as age with polypharmacy and comorbidities[33], as well as specific drug-comorbidity, drug-drug and demographics-drug interactions (e.g., dementia with antipsychotic drugs[34–36], non-steroidal anti-inflammatory drugs (NSAIDs) with opioid analgesics[37,38], and frailty with antihistamines[39,40]). Beyond confirming existing knowledge, our framework also identified some novel hypotheses that deserve further exploration. For example, the system identified a non-linear relationship between alcohol status and FRI risk (supported by recent literature [41]) and a compounded risk interaction between dementia and antispasmodic medications (can be explained as the anticholinergic effects carried by antispasmodics can result in impairment of memory function[42], mirroring the symptoms of dementia). Crucially, the Recommender supports rather than replaces clinical judgment. For instance, while the tool suggested excluding visual impairment, likely due to low frequency/predictive power, the framework's transparent design allows researchers to critically evaluate such data-driven recommendations.

This work extends previous research to predictive model designs in several ways. First, the framework is not simply a feature selection tool. Existing approaches, including robust variable selection and regularization-based methods for Cox models [43], provide rigorous strategies for deriving effective predictor sets, improving model stability, and reducing overfitting within prespecified modelling frameworks. However, these methods are primarily designed to determine which variables should be retained in a final model. In contrast, our framework combines a flexible non-linear exploratory model with explainable AI to generate broader design-relevant insights, including candidate feature exclusions, potential non-linear effects, interaction structures, and subgroup-specific predictor behaviour. These outputs are intended not only to support model development, but also to help researchers understand why a feature may be informative, how its contribution may vary across patient subgroups, and which feature relationships may warrant formal statistical evaluation. In this sense, the framework complements rather than replaces established biostatistical methods by prioritizing clinically and statistically plausible signals from a large candidate space for targeted follow-up using conventional tools, including penalized Cox models, restricted cubic splines, interaction terms, and likelihood ratio tests.

Second, the Recommender provides a mechanism for exploring subgroup-specific predictor behaviour that may be overlooked by conventional model development. Traditional models often emphasize high-risk populations, yet patients classified as low risk are not necessarily free from risk; rather, potential risk factors may be compensated by other protective factors.[44] Our extreme-group analysis revealed that NSAIDs, although contributing minimally in the high-risk cohort, was a significant predictor specifically in low-risk cohorts. Identifying these predictors in low-risk groups offers new opportunities for early, targeted preventive interventions. Furthermore, this stratified mechanism offers visibility into how clinical features behave differently across distinct subpopulations: atrial fibrillation generally decreased predicted risk for patients under 65 years but increased it among those aged 65 years or older (Figure 5). While the framework does not automatically calculate formal mathematical fairness metrics, its architecture actively facilitates algorithmic equity audits. Researchers can define strata using protected characteristics such as sex, ethnicity, or socioeconomic deprivation. The system then systematically isolates and compares variable importance across these demographic subsets. By exposing these subgroup-specific deviations, the framework provides the exact insights required to identify bias early in the study design.

Third, this framework bridges the gap between AI power and clinical transparency. AI has demonstrated remarkable predictive capabilities with complex and large-scale biomedical data,[45] but their opacity often prevents trust and clinical adoption.[46] By using AI models purely as an *exploratory tool* to inform conventional statistical models, we address the accuracy-interpretability trade-off without the fidelity issues of post-hoc explainable AI. [47] The exploratory RSF achieving a higher C-index than the augmented Cox model (0.820 vs. 0.815) is a good example of this as the latter provides a critical advantage in highly regulated clinical environments where a fixed, transparent formula is often preferred over an ensemble of trees for safety and auditability.

Furthermore, our framework demonstrates some utility in a "hard-to-improve" scenario as the improvement in C-index (0.010) is statistically significant but numerically modest. Relatively simple models can perform reasonably well on high-quality, large-scale datasets, leading to great difficulty in achieving marginal gains. However, in large-scale population health, even marginal gains in discrimination and calibration can lead to significantly better resource allocation and risk stratification when applied to hundreds of thousands of patients. Our approach is: scalable as it rapidly explores thousands of pairwise interactions; efficient as it avoids the computational demands of complex architectures like Graph Neural Networks;[48,49] adaptable as it is applicable to different models (e.g., DeepSurv[50] or diagnosis prediction) and clinical datasets. While SHAP is utilized here due to its theoretical foundations in game theory, the framework is attribution-agnostic. Researchers may consider other techniques like Integrated Gradients[51] or SurvSHAP(t)[52] for time-dependent explanations, depending on the computational budget and the specific exploratory model used.

This framework offers practical implications for clinical research. It streamlines future data collection by reducing variable burden without sacrificing predictive performance, can guide feature selection in further causal studies, and enables researchers to move beyond preconceived hypotheses.[53] More broadly, the framework encourages researchers to move beyond purely literature-driven or knowledge-based model specification by providing a systematic, data-driven mechanism for discovering important features, non-linearities, interactions, and subgroup-specific effects. These insights may also inform subsequent causal studies, although they should not be interpreted as causal evidence without further causal design and validation.

This study has several limitations. First, our interaction terms were restricted to (already challenging) first-order pairs to ensure computational feasibility. Higher-order interactions were not explored, but are a potential focus for future research, where computational complexity is the main concern.[54] Second, the validity of the recommendations depends on the exploratory model's performance (e.g., RSF). Suboptimal discrimination or calibration could introduce bias into FAs, necessitating careful model selection or ensemble approaches. Thus, we advise using the recommendations produced only with high performing exploratory models (high C-index and stable calibration), and leave investigation into quality metrics of FAs to future work. Third, our analysis focused on large-scale populations rather than individual longitudinal trajectories, leaving temporal dependencies out of scope, though in principle applicable with extensions.[52] Fourth, we applied recommendations in a fixed sequence, starting with feature exclusion to reduce complexity, which may yield different results than alternative ordering. Fifth, the baseline Cox model was intentionally kept 'naïve' to represent standard literature-driven practice. We acknowledge that penalized models like LASSO or Elastic Net[55] are strong benchmarks. While our current study focuses on the discovery of interactions and non-linearities via explainable AI, future work should benchmark this framework against these established regularization techniques to quantify the gain of explainable AI-driven design. Finally, we emphasize that FAs represent predictive associations, not causal relationships; they denote contributions to the model's output rather than the clinical outcome itself. Any clinical interventions based on these findings would require subsequent causal discovery or randomized controlled trials.

# Conclusion

We demonstrated that explainable AI can uncover complex feature relationships within high-dimensional clinical datasets. This data-driven exploratory approach showed the potential to improve the robustness of predictive cohort study design by augmenting traditional, expert-led methods. These findings also suggest that explainable AI-informed insights can generate novel hypotheses that may warrant further investigation.

## Data Availability

The data that support the findings of this study are available from DataLoch (https://dataloch.org/), but restrictions apply to the availability of these data, which were used under license for the current study, and so are not publicly available. DataLoch data are available to researchers who complete governance training and obtain approvals. Access to DataLoch requires submission of a proposal, with investigator support and approval as needed. DataLoch data is stored in an NHS safe haven and can be accessed remotely. GBSG2 and ACT are publicly available.

## Code Availability

The code to replicate the analysis on publicly available datasets (GBSG2 and ACT) is available on GitHub via https://github.com/CHAI-UK/XAI-insight-discovery/.

# Acknowledgements

We acknowledge the support of the UKRI AI programme, and the Engineering and Physical Sciences Research Council (EPSRC), for the Causality in Healthcare AI Hub [grant number EP/Y028856/1]. Junyu Yan is supported in part by the Advanced Care Research Center by the Ph.D. studentship.

**Figure 1**. Overview of the incorporated methodology.

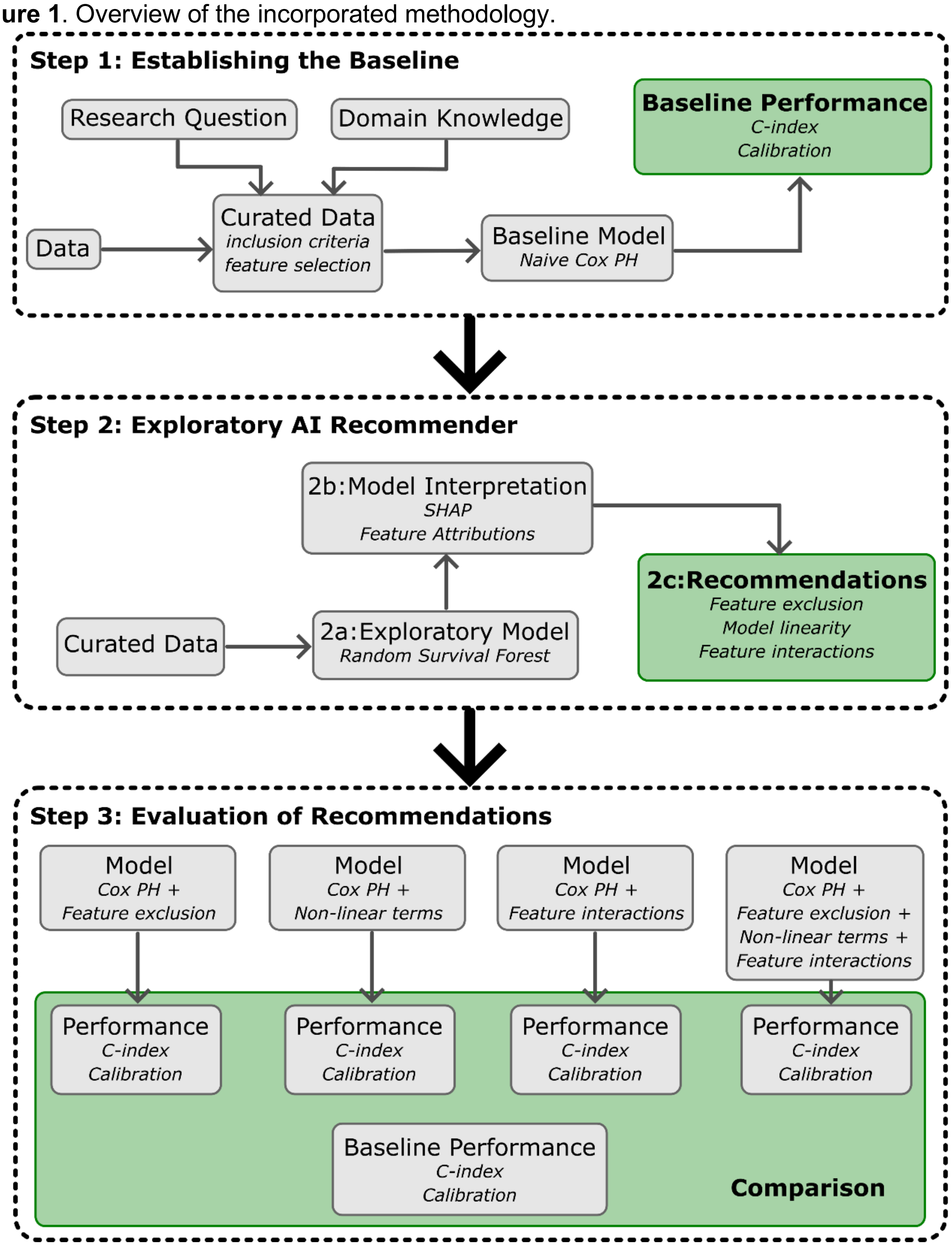

The incorporated methodology consists of three main steps. *Step 1* intends to mimic the usual knowledge-driven practice without any AI-based tools. The raw *Data* is curated by the research team in accordance with their *Research Question* and using their *Domain Knowledge*. The resulting *Curated Data* is then used with a naive Cox model to obtain the *Baseline Performance. Step 2* demonstrates the use of our proposed *Exploratory AI Recommender*, a data-driven framework to obtain feature-related recommendations. *Curated Data*, prepared in the previous step, is fed into the *Exploratory Model (Step 2a)* that uses flexible AI methods (e.g., *Random Survival Forest*) to capture complex data patterns. *Model Interpretation* techniques (*Step 2b*), such as *Feature Attributions*, are then applied to extract those patterns from the *Exploratory Model*. Next, *Feature Attributions* are processed with a pre-defined set of rules (see Figure 2) to obtain feature-related *Recommendations* (*Step 2c*). In *Step 3*, we evaluate the effectiveness of the provided recommendations. To achieve this, we take the *Baseline Model* from *Step 1* and extend it with recommendations from *Step 2c*. The performances of augmented models are then compared to the performance of the baseline model using *C-index* and *Calibration* plots.

**Figure 2.** Step 2c of our proposed methodology providing feature-related recommendations.

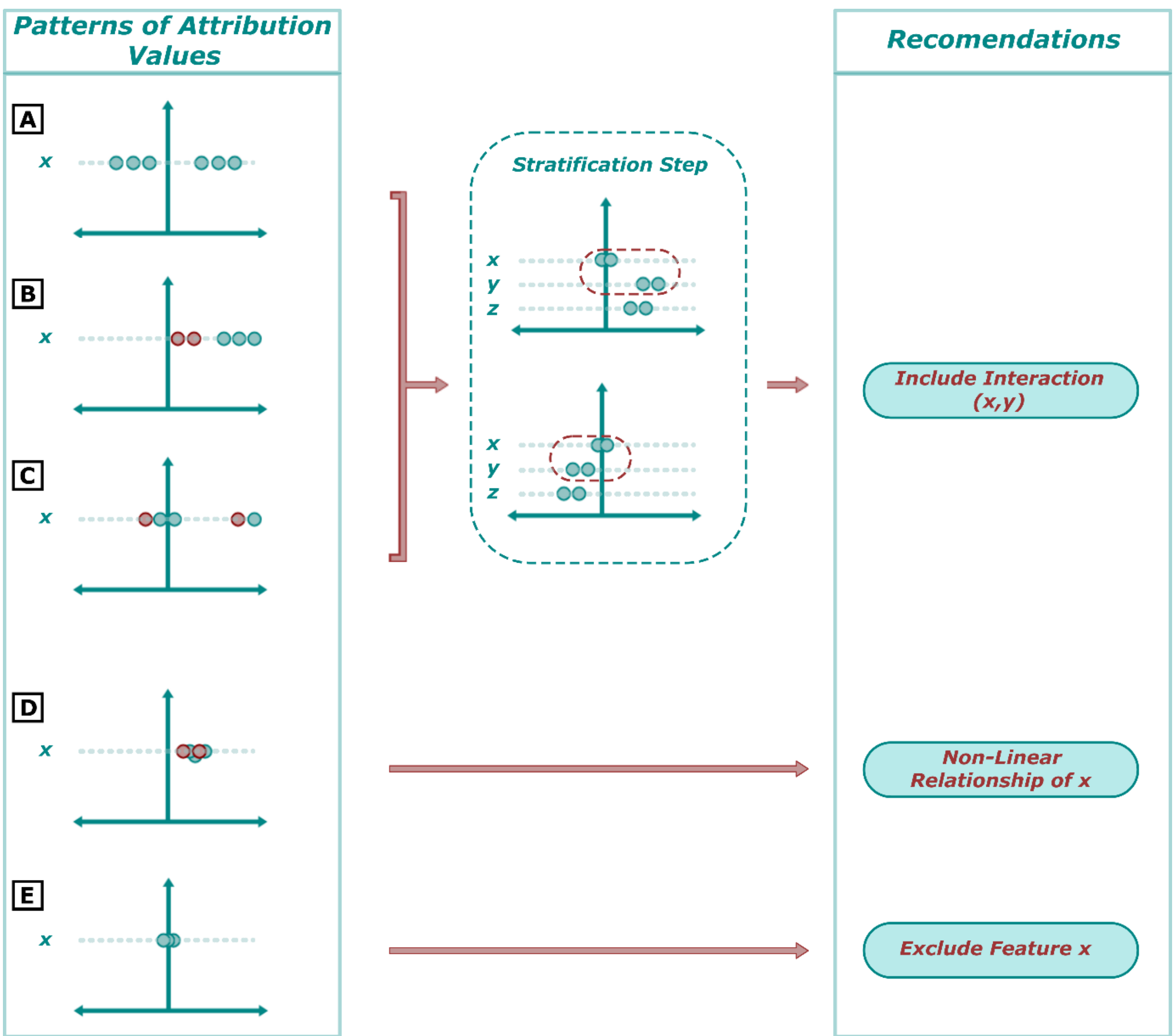

There are five patterns of feature attribution values to be recognized (A-E), based on which our framework makes three types of *Recommendations*. Three of the patterns (A-C) are associated with stratification and involve observing groups of points that can be separated based on: A) feature attribution sign, B) feature values, and C) feature attribution values. Pattern A) fits when positive and negative feature attributions form distinct groups. Pattern B) can be recognized when separate groups consist of different feature values. Pattern C) may involve mixed feature values, but the distinct groups are identified based on separation across feature attribution values. Once the groups are identified, feature attributions are created for each stratum, which notably includes finding a new reference point for each group (feature-average within each stratum). The values for both strata are analyzed for differences in other (not stratified on) features (y and z in the figure). If a feature's distributions are statistically different between the two strata (e.g., feature y) then our framework recommends including an interaction between that feature (y) and the feature we stratified on (x). The other two patterns (D-E) do not involve stratification. If feature attributions (x-axis) and feature values (colors) are mixed but not centered around and not close to zero (pattern D), it suggests the feature influences the model in a non-linear way, which is why our framework recommends creating a non-linear term (e.g., quadratic) for that feature moving forward. The fifth and last pattern involves detecting feature attributions that are mostly centered around the value of zero (pattern E). If this is the case, it suggests that the feature has negligible or no impact on model predictions, leading to recommended exclusion of the feature from further analysis.

**Figure 3.** Evaluation of the recommendations provided by the proposed framework as compared to the naïve baseline model.

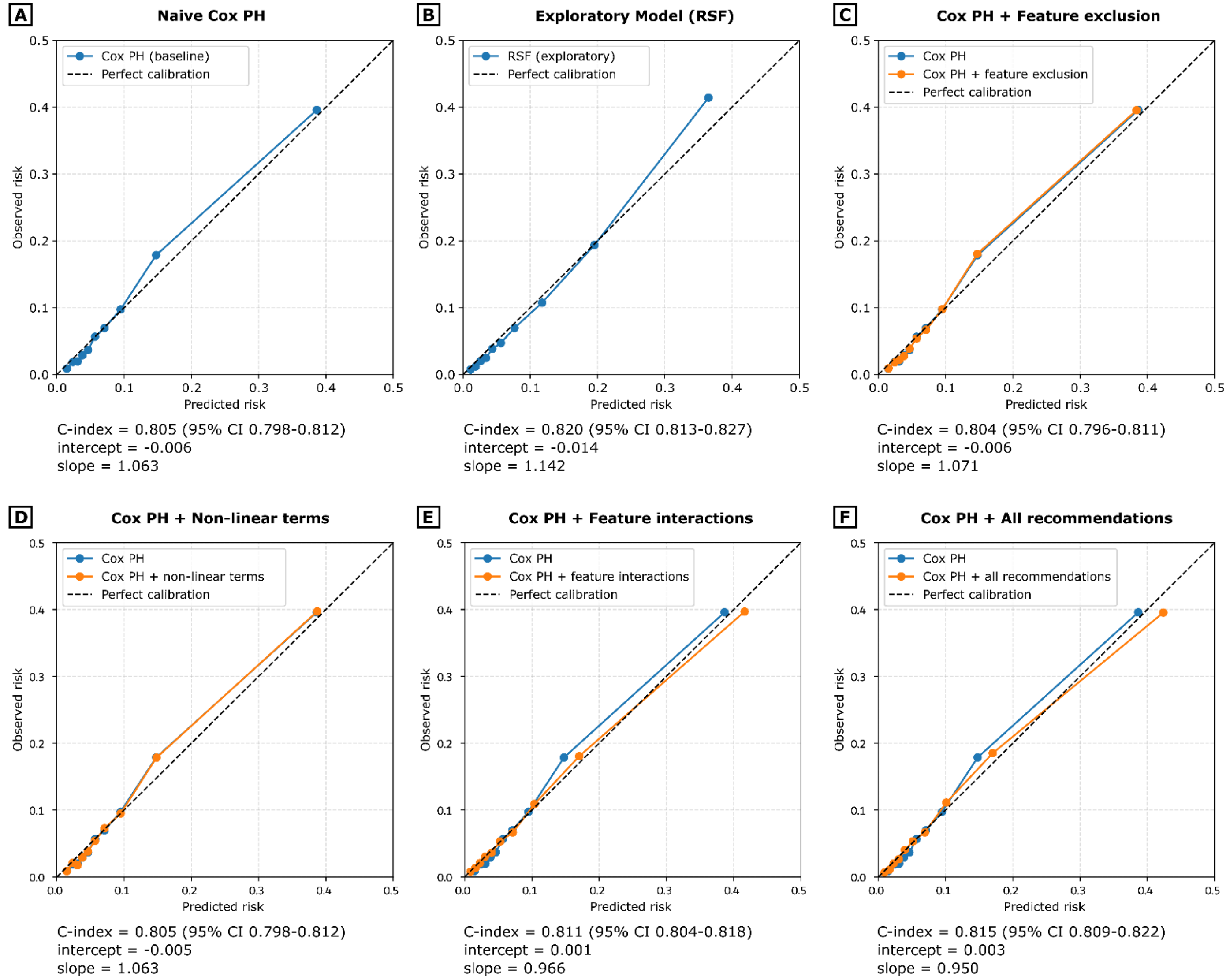

**Figure 4.** Feature attributions from Random Survival Forest estimating FRI risks among individuals within low-risk and high-risk cohorts.

Feature attributions for low-risk and high-risk groups. Each dot corresponds to a single patient observation, and its position along x-axis indicates its influence on the risk of an FRI event. Positive FAs (right of 0) indicate that a feature increased the FRI risk, whereas negative FAs (left of 0) indicate that a feature is correlated with decreased FRI risk. Color is used to indicate the value of included feature characteristics. Red and blue indicate higher and lower feature values, respectively. Medications and comorbidities are binary features hence red indicates presence of the listed characteristic (e.g., diagnosis of *angina*) and blue its absence (e.g., no diagnosis of *angina*). Marked data points (dashed lines) showcase examples of how the pre-defined patterns (see Figure 2) are being processed into recommendations in practice.

**Figure 5.** Age-stratified feature attribution summary among low-risk individuals with age lower and greater than 65.

Feature attributions for two age groups. Patients were stratified into lower-age group (n=1023) and higher-age group (n=536). The low-age group included individuals whose predicted risk score was within a predefined score margin based on an average low-risk patient with age below 65 in the test data (feature averaged across all patients in the test data with no records of falls and age below 65). This margin was defined as a Euclidean distance less than 0.01 multiplied by the standard deviation of the model's predictions across the test set. The high-age cohort included individuals whose predicted risk score was within the same score margin but whose age is above 65. Marked data points (dashed lines) show examples of patterns that lead to recommended interactions. Specifically, marked points show different distributions in the two age strata, which is a sign of potential interactions with the *age* feature

**Figure 6.** Results for the GBSG2 dataset.

**A**

POPULATION: LOW-RISK

POPULATION: HIGH-RISK

Feature Attribution

**B**

**Naive Cox PH**

C-index = 0.665 (95% CI 0.665-0.665)
intercept = 0.106
slope = 0.963

**Exploratory Model (RSF)**

C-index = 0.673 (95% CI 0.673-0.673)
intercept = -0.004
slope = 1.163

**Cox PH + All recommendations**

C-index = 0.687 (95% CI 0.687-0.687)
intercept = 0.112
slope = 0.945

A (top): Feature attributions obtained on the low-risk (left) and high-risk (right) populations. B (bottom): Calibration plots and predictive performances (C-index) of the baseline Cox PH model, the RSF exploratory model, and the Cox PH model augmented with all recommendations.

**Figure 7.** Results for the ACT dataset.

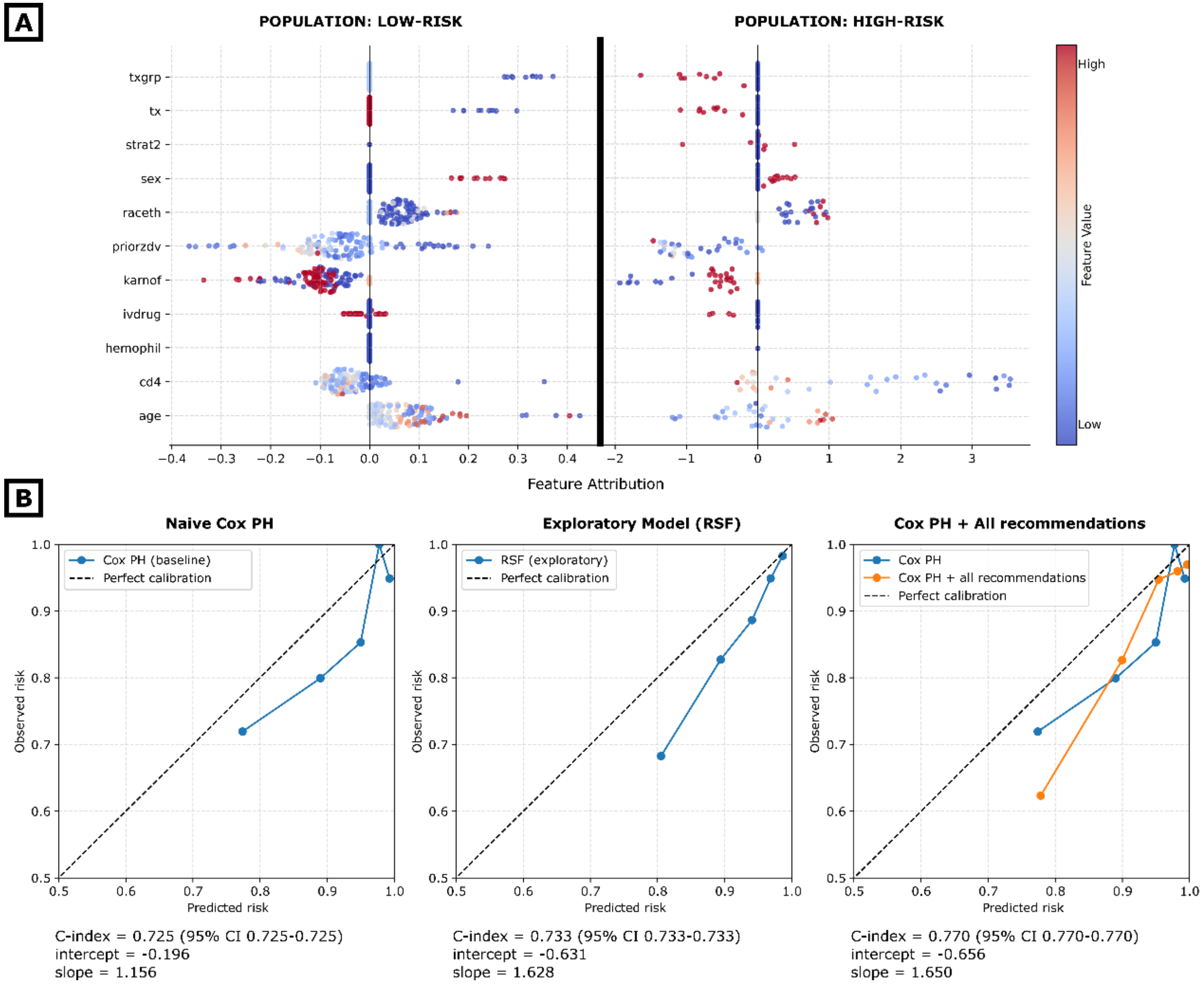

A (top): Feature attributions obtained on the low-risk (left) and high-risk (right) populations. B (bottom): Calibration plots and predictive performances (C-index) of the baseline Cox PH model, the RSF exploratory model, and the Cox PH model augmented with all recommendations.

# Supplementary Information

**Supplementary Table 1: International Classification of Diseases Clinical Codes for Fall and Related Injuries. Falls were identified based on 72 Read version 2 (Read2) codes (ascertaining falls in GP records), 89 ICD-10 codes (ascertaining falls in hospital discharge diagnoses), and three presenting complaint codes (ascertaining falls in emergency department [ED] data). Relevant fractures were identified using 78 GP Read2 codes, seven ICD-10 codes and seven ED diagnosis discharge codes**

| Type of FRI | Code Type | Diagnosis Codes |
|---|---|---|
| Falls | Read Codes | 16D..00, 16D6.00, 8BIG.00, R200.12, TC…00, TC0..00, TC00.00, TC00000, TC01.00, TC01000, TC01100, TC01z00, TC02.00, TC02000, TC02100, TC02z00, T04..00, TC0z.00, TC…11, TC42.00, TC42000, TC42z00, TC5..00, TC50.00, TC51.00, U10..00, U101400, U101z00, U107.00, U107z00, U108000, U10A.00, U10A511, U10Az00, U10J.00, U10J000, U10z000, U10z.00, U10zz00, U10z400, T240100, TCy..00, TCy0.00, U105000, U105.00, U107000, TCz..00, TCz.00, T040100, U101100, TC50.00, U10z.00, T040.00, TC42100, Tcyz.00, TC52.00, TC7..00, EMINSNQPA3, TC5..99, ^ESCTEL585171, HNGNQRF118, U101.00, TC5z.00, U108z00, U101000, ^ESCTUN670525, 16D1.00, TC0..99, TC52.00, TCz..99, U10A000 |
| | ICD-10 | W01X, W010, W011, W012, W013, W014, W015, W016, W017, W018, W019, W05X, W050, W051, W052, W053, W054, W055, W056, W057, W058, W059, W06X, W060, W061, W062, W063, W064, W065, W066, W067, W068, W069, W70X, W070, W071, W072, W073, W074, W075, W076, W077, W078, W079, W08X, W080, W081, W082, W083, W084, W085, W086, W087, W088, W089, W10X, W100, W101, W102, W103, W104, W105, W106, W107, W108, W109, W18X, W180, W181, W182, W183, W184, W185, W186, W187, W188, W189, W19X, W190, W191, W192, W193, W194, W195, W196, W197, W198, W199, R296 |
| | Presenting complaint | Fall, unsteady on feet, unsteady on feet/falls/longlie |
| Fracture | Read Codes | S300y11, S300z00, S302.00, S302000, S302100, S302200, S302300, S302400, S302z00, S304.00, S305.00, S30y.00, S30y.11, S30..00, S30..11, S300.00, S300000, S300100, S300200, S300300, S300311, S300400, S300500, S300600, S300700, S300800, S300900, S31z.00, 7K1LM00, S234.00, S234.11, S234000, S234100, S234111, S234200, S234600, S234700, S234800, S234900, S234911, S234912, S234A00, S234A11, S234A12, S234C00, S234D00, S234E00, S234F00, S234G00, S234z00, S23B.00, S23C.00, S23x000, S23x100, S23x111, S4C0000, S4C0100, S4C2000, S4C2100, S234400, S4C2.00, S234500, S4C0.00, 7K1LF00, S22..00, S220.00, S220000, S220100, S220200, S220300, S220400, S220500, S220600, S220700, S220z00, S222000, S226.00, S22z.00 |

| Type of FRI | Code Type | Diagnosis Codes |
| --- | --- | --- |
| | **ICD-10** | S42.2, S52.5, S52.6, S62.1, S72.0, S72.1, S72.2 |
| | **Diagnosis Discharge** | BBS520.2, BBS522, BBS523.1, BBS523.2, BBS524, BBS720, BBS423 |

**Supplementary Table 2: List of 27 high-level long-term conditions summarised from HDRUK CALIBER phenotypes combining primary care and hospital records. The number of phenotypes combined for each long-term condition is shown.**

| Condition | Number of CALIBER phenotypes combined |
|---|---|
| Angina | 12 |
| Anxiety disorder | 2 |
| Asthma | 2 |
| Atrial fibrillation | 8 |
| Cancers | 66 |
| Coronary | 3 |
| Dementia | 2 |
| Depression | 2 |
| Diabetes mellitus | 6 |
| End stage renal disease | 2 |
| Epilepsy | 2 |
| Gout | 2 |
| Heart failure | 2 |
| Hypertension | 3 |
| Hyperthyroidism | 2 |
| Multiple sclerosis | 1 |
| Multiple valve dz | 1 |
| Myocardial infarction | 7 |
| Osteoporosis | 2 |
| Parkinson's disease | 2 |
| Peripheral arterial disease | 5 |
| Pulmonary circulation disorders | 5 |
| Rheumatoid arthritis | 2 |
| Schizophrenia, schizotypal and delusional disorders | 2 |
| Stroke | 6 |
| Transient ischaemic attack | 2 |
| Visual impairment and blindness | 2 |

**Supplementary Table 3: Detailed list of HDRUK CALIBER phenotypes that are combined into the 27 high-level LTC groups listed into Table S2. Each CALIBER phenotype and subcategory may be mapped to multiple GP (Read) or Hospital (ICD-10) codes as per published codelists on the HDRUK Phenotype Library.**

| High-Level LTC Group | CALIBER Phenotype Name | CALIBER Phenotype Subcategory |
|---|---|---|
| Angina | Stable angina | CABG Performed (2) |
| Angina | Stable angina | Chest pain, attributed to coronary causes (4) |
| Angina | Stable angina | History of stable angina (1) |
| Angina | Stable angina | PCI performed (2) |
| Angina | Stable angina | Results abnormal (3) |
| Angina | Stable angina | Results abnormal – indicative of ischaemia (T-wave changes or ST-depression) (2) |
| Angina | Stable angina | Stable angina (4) |
| Angina | Stable angina | Stable angina admission (4) |
| Angina | Unstable angina | ACS diagnosed (3) |
| Angina | Unstable angina | Unstable angina (3) |
| Angina | Unstable angina | Worsening angina (2) |
| Angina | Unstable angina | yes (1) |
| Anxiety disorders | Anxiety disorders | Diagnosis of Anxiety disorders |
| Anxiety disorders | Anxiety disorders | History of Anxiety disorders |
| Asthma | Asthma | Diagnosis of Asthma |
| Asthma | Asthma | History of Asthma |
| Atrial Fibrillation | Atrial fibrillation | Atrial fibrillation monitoring (2) |
| Atrial Fibrillation | Atrial fibrillation | Atrial fibrillation or flutter (6) |
| Atrial Fibrillation | Atrial fibrillation | Atrial fibrillation, not otherwise specified (5) |
| Atrial Fibrillation | Atrial fibrillation | Atrial flutter (7) |
| Atrial Fibrillation | Atrial fibrillation | Fibrillation and flutter |
| Atrial Fibrillation | Atrial fibrillation | History of atrial fibrillation or flutter (1) |
| Atrial Fibrillation | Atrial fibrillation | Paroxysmal atrial fibrillation (3) |
| Atrial Fibrillation | Atrial fibrillation | Persistent or permanent atrial fibrillation (4) |
| Cancers | Hodgkin Lymphoma | Diagnosis of Hodgkin Lymphoma |
| Cancers | Hodgkin Lymphoma | History of Hodgkin Lymphoma |
| Cancers | Leukaemia | Diagnosis of Leukaemia |
| Cancers | Leukaemia | History of Leukaemia |
| Cancers | Monoclonal gammopathy of undetermined significance (MGUS) | 1 |
| Cancers | Monoclonal gammopathy of undetermined significance (MGUS) | Diagnosis of Monoclonal gammopathy of undetermined significance (MGUS) |
| Cancers | Multiple myeloma and malignant plasma cell neoplasms | Diagnosis of Multiple myeloma and malignant plasma cell neoplasms |
| Cancers | Myelodysplastic syndromes | Diagnosis of Myelodysplastic syndromes |
| Cancers | Non-Hodgkin Lymphoma | Diagnosis of Non-Hodgkin Lymphoma |
| Cancers | Polycythaemia vera | Diagnosis of Polycythaemia vera |
| Cancers | Primary Malignancy_biliary tract | Diagnosis of Primary Malignancy_biliary tract |
| Cancers | Primary Malignancy_Bladder | Diagnosis of Primary Malignancy_Bladder |
| Cancers | Primary Malignancy_Bladder | History of Primary Malignancy_Bladder |
| Cancers | Primary Malignancy_Bone and articular cartilage | Diagnosis of Primary Malignancy_Bone and articular cartilage |
| Cancers | Primary Malignancy_Bone and articular cartilage | History of Primary Malignancy_Bone and articular cartilage |
| Cancers | Primary Malignancy_Brain, Other CNS and Intracranial | Diagnosis of Primary Malignancy_Brain, Other CNS and Intracranial |
| Cancers | Primary Malignancy_Brain, Other CNS and Intracranial | History of Primary Malignancy_Brain, Other CNS and Intracranial |
| Cancers | Primary Malignancy_Breast | Diagnosis of Primary Malignancy_Breast |
| Cancers | Primary Malignancy_Breast | History of Primary Malignancy_Breast |
| Cancers | Primary Malignancy_Cervical | Diagnosis of Primary Malignancy_Cervical |
| Cancers | Primary Malignancy_Cervical | History of Primary Malignancy_Cervical |
| Cancers | Primary Malignancy_colorectal and anus | Diagnosis of Primary Malignancy_colorectal and anus |
| Cancers | Primary Malignancy_colorectal and anus | History of Primary Malignancy_colorectal and anus |
| Cancers | Primary Malignancy_Kidney and Ureter | Diagnosis of Primary Malignancy_Kidney and Ureter |
| Cancers | Primary Malignancy_Kidney and Ureter | History of Primary Malignancy_Kidney and Ureter |
| Cancers | Primary Malignancy_Liver | Diagnosis of Primary Malignancy_Liver |
| Cancers | Primary Malignancy_Liver | History of Primary Malignancy_Liver |
| Cancers | Primary Malignancy_Lung and trachea | Diagnosis of Primary Malignancy_Lung and trachea |
| Cancers | Primary Malignancy_Lung and trachea | History of Primary Malignancy_Lung and trachea |
| Cancers | Primary Malignancy_Malignant Melanoma | Diagnosis of Primary Malignancy_Malignant Melanoma |
| Cancers | Primary Malignancy_Malignant Melanoma | History of Primary Malignancy_Malignant Melanoma |
| Cancers | Primary Malignancy_Mesothelioma | Diagnosis of Primary Malignancy_Mesothelioma |
| Cancers | Primary Malignancy_Mesothelioma | History of Primary Malignancy_Mesothelioma |
| Cancers | Primary Malignancy_Multiple independent sites | Diagnosis of Primary Malignancy_Multiple independent sites |
| Cancers | Primary Malignancy_Multiple independent sites | History of Primary Malignancy_Multiple independent sites |
| Cancers | Primary Malignancy_Oesophageal | Diagnosis of Primary Malignancy_Oesophageal |
| Cancers | Primary Malignancy_Oesophageal | History of Primary Malignancy_Oesophageal |
| Cancers | Primary Malignancy_Oro-pharyngeal | Diagnosis of Primary Malignancy_Oro-pharyngeal |
| Cancers | Primary Malignancy_Oro-pharyngeal | History of Primary Malignancy_Oro-pharyngeal |
| Cancers | Primary Malignancy_Other Organs | Diagnosis of Primary Malignancy_Other Organs |
| Cancers | Primary Malignancy_Other Organs | Possible Diagnosis of Primary Malignancy_Other Organs |
| Cancers | Primary Malignancy_Ovarian | Diagnosis of Primary Malignancy_Ovarian |

| Cancers | Primary Malignancy_Ovarian | History of Primary Malignancy_Ovarian |
|---|---|---|
| Cancers | Primary Malignancy_Pancreatic | Diagnosis of Primary Malignancy_Pancreatic |
| Cancers | Primary Malignancy_Prostate | Diagnosis of Primary Malignancy_Prostate |
| Cancers | Primary Malignancy_Prostate | History of Primary Malignancy_Prostate |
| Cancers | Primary Malignancy_Prostate | Procedure for Primary Malignancy_Prostate |
| Cancers | Primary Malignancy_Stomach | Diagnosis of Primary Malignancy_Stomach |
| Cancers | Primary Malignancy_Stomach | History of Primary Malignancy_Stomach |
| Cancers | Primary Malignancy_Testicular | Diagnosis of Primary Malignancy_Testicular |
| Cancers | Primary Malignancy_Testicular | History of Primary Malignancy_Testicular |
| Cancers | Primary Malignancy_Thyroid | Diagnosis of Primary Malignancy_Thyroid |
| Cancers | Primary Malignancy_Thyroid | History of Primary Malignancy_Thyroid |
| Cancers | Primary Malignancy_Uterine | Diagnosis of Primary Malignancy_Uterine |
| Cancers | Primary Malignancy_Uterine | History of Primary Malignancy_Uterine |
| Cancers | Secondary Malignancy_Adrenal gland | Diagnosis of Secondary Malignancy_Adrenal gland |
| Cancers | Secondary Malignancy_Bone | Diagnosis of Secondary Malignancy_Bone |
| Cancers | Secondary Malignancy_Bowel | Diagnosis of Secondary Malignancy_Bowel |
| Cancers | Secondary Malignancy_Brain, Other CNS and Intracranial | Diagnosis of Secondary Malignancy_Brain, Other CNS and Intracranial |
| Cancers | Secondary malignancy_Liver and intrahepatic bile duct | Diagnosis of Secondary malignancy_Liver and intrahepatic bile duct |
| Cancers | Secondary Malignancy_Lung | Diagnosis of Secondary Malignancy_Lung |
| Cancers | Secondary Malignancy_Lymph Nodes | Diagnosis of Secondary Malignancy_Lymph Nodes |
| Cancers | Secondary Malignancy_Other organs | Diagnosis of Secondary Malignancy_Other organs |
| Cancers | Secondary Malignancy_Other organs | Possible diagnosis of Secondary Malignancy_Other organs |
| Cancers | Secondary Malignancy_Pleura | Diagnosis of Secondary Malignancy_Pleura |
| Cancers | Secondary Malignancy_retroperitoneum and peritoneum | Diagnosis of Secondary Malignancy_retroperitoneum and peritoneum |
| Coronary | Coronary heart disease not otherwise specified | Diagnosed |
| Coronary | Coronary heart disease not otherwise specified | Diagnosed (3) |
| Coronary | Coronary heart disease not otherwise specified | History of (1) |
| Dementia | Dementia | Diagnosis of Dementia |
| Dementia | Dementia | History of Dementia |
| Depression | Depression | Diagnosis of Depression |
| Depression | Depression | History of Depression |
| Diabetes Mellitus | Diabetes | Diabetes not otherwise specified (6) |
| Diabetes Mellitus | Diabetes | Insulin dependent diabetes (3) |
| Diabetes Mellitus | Diabetes | Non insulin dependent diabetes (4) |

| Diabetes Mellitus | Diabetes | Secondary diabetes (5) |
|---|---|---|
| Diabetes Mellitus | Diabetes | Type I diabetes mellitus (3) |
| Diabetes Mellitus | Diabetes | Type II diabetes mellitus (4) |
| End Stage Renal Disease | End Stage Renal Disease | Diagnosis of End stage renal disease |
| End Stage Renal Disease | End Stage Renal Disease | Procedure for End stage renal disease |
| Epilepsy | Epilepsy | Diagnosis of Epilepsy |
| Epilepsy | Epilepsy | History of Epilepsy |
| Gout | Gout | Diagnosis of Gout |
| Gout | Gout | History of Gout |
| Heart Failure | Heart failure | Diagnosis of Heart failure |
| Heart Failure | Heart failure | History of Heart failure |
| Hypertension | Hypertension | History of hypertension (1) |
| Hypertension | Hypertension | Hypertension (3) |
| Hypertension | Hypertension | Secondary hypertension (4) |
| Hyperthyroidism | Hypo or hyperthyroidism | Diagnosis of Hypo or hyperthyroidism |
| Hyperthyroidism | Hypo or hyperthyroidism | History of Hypo or hyperthyroidism |
| Multiple Sclerosis | Liver fibrosis, sclerosis and cirrhosis | Diagnosis of Liver fibrosis, sclerosis and cirrhosis |
| Multiple Valve dz | Multiple valve dz | Diagnosis of Multiple valve dz |
| Myocardial Infarction | Myocardial infarction | Acute MI not further specified (5) |
| Myocardial Infarction | Myocardial infarction | Complications of MI (7) |
| Myocardial Infarction | Myocardial infarction | History of MI (1) |
| Myocardial Infarction | Myocardial infarction | NSTEMI (4) |
| Myocardial Infarction | Myocardial infarction | Possible MI or uncertain date (2) |
| Myocardial Infarction | Myocardial infarction | STEMI (3) |
| Myocardial Infarction | Myocardial infarction | Subsequent MI (6) |
| Osteoporosis | Osteoporosis | Diagnosis of Osteoporosis |
| Osteoporosis | Osteoporosis | History of Osteoporosis |
| Parkinson's Disease | Parkinson's disease | Diagnosis of Parkinson's disease |
| Parkinson's Disease | Parkinson's disease | History of Parkinson's disease |
| Peripheral Arterial Disease | Peripheral arterial disease | history of PVD during a consultation |
| Peripheral Arterial Disease | Peripheral arterial disease | Leg or aortic embolism or thrombosis (8) |
| Peripheral Arterial Disease | Peripheral arterial disease | Other PAD procedures (3) |
| Peripheral Arterial Disease | Peripheral arterial disease | PAD procedure performed (3) |
| Peripheral Arterial Disease | Peripheral arterial disease | Peripheral vascular disease (7) |
| Pulmonary Circulation Disorders | Primary pulmonary hypertension | Diagnosis of Primary pulmonary hypertension |
| Pulmonary Circulation Disorders | Pulmonary embolism | Diagnosis of Pulmonary embolism |
| Pulmonary Circulation Disorders | Pulmonary embolism | History of Pulmonary embolism |
| Pulmonary Circulation Disorders | Pulmonary embolism | Procedure for Pulmonary embolism |

| Pulmonary Circulation Disorders | Secondary pulmonary hypertension | Diagnosis of Secondary pulmonary hypertension |
|---|---|---|
| Rheumatoid Arthritis | Rheumatoid Arthritis | Diagnosis of Rheumatoid Arthritis |
| Rheumatoid Arthritis | Rheumatoid Arthritis | History of Rheumatoid Arthritis |
| Schizophrenia, schizotypal and delusional disorders | Schizophrenia, schizotypal and delusional disorders | Diagnosis of Schizophrenia, schizotypal and delusional disorders |
| Schizophrenia, schizotypal and delusional disorders | Schizophrenia, schizotypal and delusional disorders | History of Schizophrenia, schizotypal and delusional disorders |
| Stroke | Ischaemic stroke | Diagnosis of Ischaemic stroke |
| Stroke | Ischaemic stroke | History of cerebral infarction |
| Stroke | Ischaemic stroke | History of Ischaemic stroke |
| Stroke | Stroke NOS | Diagnosis of Stroke NOS |
| Stroke | Stroke NOS | History of Stroke NOS |
| Stroke | Stroke NOS | History of stroke, not specified as haemorrhage or infarction |
| Transient Ischaemic Attack | Transient ischaemic attack | Diagnosis of TIA |
| Transient Ischaemic Attack | Transient ischaemic attack | History of TIA |
| Visual Impairment and Blindness | Visual Impairment and Blindness | Diagnosis of blindness |
| Visual Impairment and Blindness | Visual Impairment and Blindness | Diagnosis of Visual impairment and blindness |

**Supplementary Table 4: Population details of the DataLoch dataset.**

| Feature | No. (%) | | | |
|---|---|---|---|---|
| | Entire cohort (N = 245614) | Non-FRI (N = 227144) | FRI (N = 18740) | P value |
| **Age, mean (SD), y** | 67.3 (11.6) | 66.7 (11.3) | 74.8 (12.2) | <0.001 |
| **SIMD quintiles** | | | | |
| **1 (least deprived)** | 33365 (13.6) | 30476 (13.4) | 2889 (15.6) | <0.001 |
| **2** | 53148 (21.6) | 48612 (21.4) | 4536 (24.6) | <0.001 |
| **3** | 42130 (17.2) | 39103 (17.2) | 3027 (16.4) | <0.001 |
| **4** | 42549 (17.3) | 39347 (17.3) | 3202 (17.3) | <0.001 |
| **5** | 74422 (30.3) | 69606 (30.6) | 4816 (26.1) | <0.001 |
| **Smoking Status** | | | | |
| **Current smoker** | 40192 (16.4) | 37222 (16.4) | 2970 (16.1) | <0.001 |
| **Ex-smoker** | 115015 (46.8) | 105870 (46.6) | 9145 (49.5) | <0.001 |
| **Missing** | 491 (0.2) | 464 (0.2) | 27 (0.1) | <0.001 |
| **Non-smoker** | 89916 (36.6) | 83588 (36.8) | 6328 (34.3) | <0.001 |
| **EFI Value, mean (SD)** | 0.1 (0.1) | 0.1 (0.1) | 0.2 (0.1) | <0.001 |
| **BMI** | | | | |
| **0** | 7876 (3.2) | 6804 (3.0) | 1072 (5.8) | <0.001 |
| **1** | 77450 (31.5) | 70437 (31.0) | 7013 (38.0) | <0.001 |
| **2** | 85741 (34.9) | 79968 (35.2) | 5773 (31.3) | <0.001 |
| **3** | 64319 (26.2) | 60305 (26.5) | 4014 (21.7) | <0.001 |
| **4** | 10228 (4.2) | 9630 (4.2) | 598 (3.2) | <0.001 |
| **Alcohol Status** | | | | |
| **Alcohol intake above. recommended sensible limits** | 9302 (3.8) | 8686 (3.8) | 616 (3.3) | <0.001 |
| **Alcohol intake within. recommended sensible limits** | 63134 (25.7) | 58752 (25.9) | 4382 (23.7) | <0.001 |
| **Drinks rarely** | 59027 (24.0) | 54749 (24.1) | 4278 (23.2) | <0.001 |
| **Heavy drinker - 7-9u/day** | 1889 (0.8) | 1741 (0.8) | 148 (0.8) | <0.001 |
| **Light drinker - 1-2u/day** | 20628 (8.4) | 19394 (8.5) | 1234 (6.7) | <0.001 |
| **Missing** | 33348 (13.6) | 31024 (13.7) | 2324 (12.6) | <0.001 |
| **Moderate drinker - 3-6u/day** | 11477 (4.7) | 10871 (4.8) | 606 (3.3) | <0.001 |
| **Non-drinker alcohol** | 45768 (18.6) | 40984 (18.0) | 4784 (25.9) | <0.001 |
| **Very heavy drinker - >9u/day** | 1041 (0.4) | 943 (0.4) | 98 (0.5) | <0.001 |
| **Ethnicity** | | | | |
| **Asian** | 3444 (1.4) | 3335 (1.5) | 109 (0.6) | <0.001 |
| **Black** | 891 (0.4) | 870 (0.4) | 21 (0.1) | <0.001 |
| **Missing** | 21248 (8.7) | 20486 (9.0) | 762 (4.1) | <0.001 |
| **Mixed** | 529 (0.2) | 514 (0.2) | 15 (0.1) | <0.001 |
| **Not Stated** | 10784 (4.4) | 10237 (4.5) | 547 (3.0) | <0.001 |
| **Other** | 1464 (0.6) | 1412 (0.6) | 52 (0.3) | <0.001 |
| **White** | 207254 (84.4) | 190290 (83.8) | 16964 (91.8) | <0.001 |
| **Sex** | | | | |
| **Female** | 114067 (46.4) | 108327 (47.7) | 5740 (31.1) | <0.001 |
| **Male** | 131547 (53.6) | 118817 (52.3) | 12730 (68.9) | <0.001 |
| **Medications** | | | | |
| **Thiazide and related diuretics** | 34967 (14.2) | 31891 (14.0) | 3076 (16.7) | <0.001 |
| **Skeletal muscle relaxants** | 2161 (0.9) | 1968 (0.9) | 193 (1.0) | 0.014 |
| **Selective serotonin re-uptake inhibitors** | 40490 (16.5) | 37230 (16.4) | 3260 (17.7) | <0.001 |
| **Other antidepressant drugs** | 21211 (8.6) | 19231 (8.5) | 1980 (10.7) | <0.001 |
| **Other antianginal drugs** | 3247 (1.3) | 2881 (1.3) | 366 (2.0) | <0.001 |
| **Opioid analgesics** | 188237 (76.6) | 174412 (76.8) | 13825 (74.9) | <0.001 |
| **Non-steroidal anti-inflammatory drugs** | 71078 (28.9) | 67806 (29.9) | 3272 (17.7) | <0.001 |
| **Nitrates** | 23614 (9.6) | 21456 (9.4) | 2158 (11.7) | <0.001 |
| **Loop diuretics** | 26056 (10.6) | 22516 (9.9) | 3540 (19.2) | <0.001 |
| **Hypnotics** | 25630 (10.4) | 23522 (10.4) | 2108 (11.4) | <0.001 |
| **Drugs used in status epilepticus** | 191 (0.1) | 174 (0.1) | 17 (0.1) | 0.557 |
| **Control of epilepsy** | 31789 (12.9) | 28972 (12.8) | 2817 (15.3) | <0.001 |
| **Cardiac glycosides** | 5632 (2.3) | 4841 (2.1) | 791 (4.3) | <0.001 |
| **Calcium-channel blockers** | 57705 (23.5) | 53056 (23.4) | 4649 (25.2) | <0.001 |
| **Beta-adrenoceptor blocking drugs** | 54345 (22.1) | 49737 (21.9) | 4608 (24.9) | <0.001 |
| **Anxiolytics** | 30572 (12.4) | 28508 (12.6) | 2064 (11.2) | <0.001 |
| **Antispasmod.&other drgs. alt.gut motility** | 24631 (10.0) | 23336 (10.3) | 1295 (7.0) | <0.001 |
| **Antipsychotic drugs** | 11933 (4.9) | 10990 (4.8) | 943 (5.1) | 0.108 |
| **Antimuscarinic drugs used in parkin'ism** | 620 (0.3) | 543 (0.2) | 77 (0.4) | <0.001 |
| **Antihistamines** | 48350 (19.7) | 45225 (19.9) | 3125 (16.9) | <0.001 |
| **Angiotensin-ii receptor antagonists** | 25941 (10.6) | 23851 (10.5) | 2090 (11.3) | <0.001 |
| **Angiotensin-converting enzyme inhibitors** | 66644 (27.1) | 61650 (27.1) | 4994 (27.0) | 0.769 |
| **Alpha-adrenoceptor blocking drugs** | 10494 (4.3) | 9543 (4.2) | 951 (5.1) | <0.001 |
| **Comorbidities** | | | | |
| **Visual impairment and. blindness** | 1320 (0.5) | 1079 (0.5) | 241 (1.3) | <0.001 |
| **Transient ischaemic attack** | 4030 (1.6) | 3632 (1.6) | 398 (2.2) | <0.001 |
| **Stroke** | 9822 (4.0) | 8715 (3.8) | 1107 (6.0) | <0.001 |
| **Schizophrenia, schizotypal and delusional disorders** | 1134 (0.5) | 1019 (0.4) | 115 (0.6) | <0.001 |
| **Rheumatoid arthritis** | 3186 (1.3) | 2882 (1.3) | 304 (1.6) | <0.001 |
| **Pulmonary circulation disorders** | 22283 (9.1) | 20173 (8.9) | 2110 (11.4) | <0.001 |
| **Peripheral arterial disease** | 4347 (1.8) | 3985 (1.8) | 362 (2.0) | 0.045 |
| **Parkinson's disease** | 1302 (0.5) | 1039 (0.5) | 263 (1.4) | <0.001 |
| **Osteoporosis** | 9487 (3.9) | 6848 (3.0) | 2639 (14.3) | <0.001 |
| **Myocardial infarction** | 8033 (3.3) | 7386 (3.3) | 647 (3.5) | 0.068 |

| **Multiple valve dz** | 1406 (0.6) | 1215 (0.5) | 191 (1.0) | <0.001 |
|---|---|---|---|---|
| **Multiple sclerosis** | 705 (0.3) | 628 (0.3) | 77 (0.4) | <0.001 |
| **Hyperthyroidism** | 14250 (5.8) | 13142 (5.8) | 1108 (6.0) | 0.240 |
| **Hypertension** | 52330 (21.3) | 48262 (21.2) | 4068 (22.0) | 0.013 |
| **Heart failure** | 8071 (3.3) | 7097 (3.1) | 974 (5.3) | <0.001 |
| **Gout** | 5374 (2.2) | 4992 (2.2) | 382 (2.1) | 0.258 |
| **Epilepsy** | 2343 (1.0) | 2049 (0.9) | 294 (1.6) | <0.001 |
| **End stage renal disease** | 911 (0.4) | 841 (0.4) | 70 (0.4) | 0.900 |
| **Diabetes mellitus** | 20529 (8.4) | 18729 (8.2) | 1800 (9.7) | <0.001 |
| **Depression** | 10896 (4.4) | 10022 (4.4) | 874 (4.7) | 0.044 |
| **Dementia** | 10225 (4.2) | 8032 (3.5) | 2193 (11.9) | <0.001 |
| **Coronary** | 20177 (8.2) | 18429 (8.1) | 1748 (9.5) | <0.001 |
| **Cancers** | 27371 (11.1) | 25484 (11.2) | 1887 (10.2) | <0.001 |
| **Atrial fibrillation** | 15717 (6.4) | 13569 (6.0) | 2148 (11.6) | <0.001 |
| **Asthma** | 11805 (4.8) | 11017 (4.9) | 788 (4.3) | <0.001 |
| **Anxiety disorder** | 7884 (3.2) | 7298 (3.2) | 586 (3.2) | 0.782 |
| **Angina** | 8078 (3.3) | 7409 (3.3) | 669 (3.6) | 0.009 |

**Supplementary Figure 1: Summary of recommendations generated by the Exploratory AI Recommender for the research question: Predict the risk of FRI.**

## Feature Exclusion

alpha-adrenoceptor blocking drugs, antimuscarinic drugs used in Parkin'ism, cardiac glycosides, drugs used in status epilepticus, other antianginal drugs, other antidepressant drugs, selective serotonin re-uptake inhibitors, skeletal muscle relaxants, angina, anxiety disorder, depression, end stage renal disease, epilepsy, gout, multiple sclerosis, multiple valve dz, myocardial infarction, parkinson's disease, rheumatoid arthritis, schizophrenia schizotypal and delusional disorders, transient ischaemic attack, visual impairment and blindness

## Model Non-linearity

Age, Alcohol

## Feature Interactions

| Stratified Variables | Interaction Variables |
|---|---|
| Age | asthma, atrial fibrillation, cancers, coronary, hypertension, pulmonary circulation disorders, stroke, alpha-adrenoceptor blocking drugs, angiotensin-ii receptor antagonists, antihistamines, antispasmod.&other drgs alt.gut motility, beta-adrenoceptor blocking drugs, calcium-channel blockers, control of epilepsy, hypnotics, loop diuretics, nitrates, non-steroidal anti-inflammatory drugs, opioid analgesics, thiazides and related diuretics, sex, ethnicity, alcohol, EFI, smoking, simd2016 quintile |
| Ethnicity | asthma, atrial fibrillation, coronary, heart failure, hypertension, pulmonary circulation disorders, stroke, angiotensin-ii receptor antagonists, antipsychotic drugs, antispasmod.&other drgs alt.gut motility, anxiolytics, calcium-channel blockers, hypnotics, opioid analgesics, thiazides and related diuretics, alcohol, BMI, EFI, smoking, simd2016 quintile |
| BMI | asthma, atrial fibrillation, cancers, coronary, diabetes mellitus, heart failure, hypertension, alpha-adrenoceptor blocking drugs, angiotensin-ii receptor antagonists, antipsychotic drugs, beta-adrenoceptor blocking drugs, calcium-channel blockers, control of epilepsy, hypnotics, nitrates, opioid analgesics, thiazides and related diuretics, sex, EFI, smoking, simd2016 quintile, age |
| EFI | asthma, cancers, coronary, diabetes mellitus, heart failure, hypertension, hyperthyroidism, pulmonary circulation disorders, alpha-adrenoceptor blocking drugs, angiotensin-ii receptor antagonists, antipsychotic drugs, antispasmod.&other drgs alt.gut motility, beta-adrenoceptor blocking drugs, calcium-channel blockers, hypnotics, loop diuretics, nitrates, opioid analgesics, sex, alcohol, simd2016_sc_quintile |
| Sex | asthma, cancers, coronary, diabetes mellitus, heart failure, hypertension, hyperthyroidism, pulmonary circulation disorders, alpha-adrenoceptor blocking drugs, angiotensin-ii receptor antagonists, antipsychotic drugs, antispasmod.&other drgs alt.gut motility, beta-adrenoceptor blocking drugs, calcium-channel blockers, hypnotics, loop diuretics, nitrates, opioid analgesics, sex, alcohol, simd2016 quintile |
| Simd2016 Quintile | diabetes mellitus, heart failure, osteoporosis, peripheral arterial disease, pulmonary circulation disorders, alpha-adrenoceptor blocking drugs, antipsychotic drugs, antispasmod.&other drgs alt.gut motility, anxiolytics, |
| Loop Diuretics | atrial fibrillation, coronary, diabetes mellitus, heart failure, hypertension, antispasmod .&other drgs alt.gut motility, anxiolytics, beta-adrenoceptor blocking drugs, calcium-channel blockers, control of epilepsy, non-steroidal anti-inflammatory drugs, opioid analgesics, thiazides and related diuretics, sex, alcohol, BMI, smoking, simd2016 quintile |
| Dementia | diabetes mellitus, hypertension, hyperthyroidism, peripheral arterial disease, pulmonary circulation disorders, antihistamines, antispasmod.&other drgs alt.gut motility, beta-adrenoceptor blocking drugs, loop diuretics, non-steroidal anti-inflammatory drugs, thiazides and related diuretics, sex, ethnicity, alcohol, BMI, age |
| Non-steroidal anti-inflammatory drugs | asthma, coronary, hyperthyroidism, alpha-adrenoceptor blocking drugs, angiotensin-ii receptor antagonists, antihistamines, antipsychotic drugs, antispasmod.&other drgs alt.gut motility, anxiolytics, beta-adrenoceptor blocking drugs, control of epilepsy, hypnotics, opioid analgesics, thiazides and related diuretics, sex, ethnicity, alcohol, BMI, EFI, smoking |
| Osteoporosis | diabetes mellitus, hypertension, antihistamines, antipsychotic drugs, antispasmod.&other drgs alt.gut motility, anxiolytics, beta-adrenoceptor blocking drugs, calcium-channel blockers, nitrates, non-steroidal anti-inflammatory drugs, opioid analgesics, sex, ethnicity, BMI, smoking, age |

**Supplementary Note 1: Feature Preprocessing**

For the baseline model, we selected 58 features comprising sociodemographic characteristics, lifestyle factors, frailty status, comorbidities, and medication use. Demographic variables included age, sex, ethnicity, and Scottish Index of Multiple Deprivation (SIMD) quintile[56]. Lifestyle risk factors included smoking status, alcohol intake, and body mass index (BMI). Frailty was assessed using the electronic Frailty Index (eFI) and categorized into 4 classes: not frail, mild frailty, moderate frailty, and severe frailty.[57] Comorbidities were determined from either GP records (Read2), or hospital (ICD-10 codes recorded in Scottish Morbidity Records [SMR]), using codelists developed by CALIBER and published in the Health Data Research (HDR) UK phenotype library.[58] To reduce redundancy and ensure focus on truly long-term health conditions, we selected 153 CALIBER phenotypes of interest after clinical review, and then grouped these into 27 long-term conditions (LTC) such as diabetes, ischaemic heart disease, and depression and related disorders (see Supplementary Table 2 and Supplementary Table 3). Medication classes of interest were defined by their British National Formulary (BNF) paragraphs. Twenty-three medication classes were selected which have previously been identified as Fall-Risk-Increasing Drugs (FRIDs),[59–61]including alpha-adrenoceptor blocking drugs, angiotensin-converting enzyme (ACE) inhibitors, angiotensin-II receptor antagonists, antihistamines, antimuscarinic drugs used in parkinsonism, antipsychotic drugs, antispasmodics and other drugs altering gut motility, anxiolytics, beta-adrenoceptor blocking drugs, calcium-channel blockers, cardiac glycosides, drugs used for the control of epilepsy or status epilepticus, hypnotics, loop diuretics, nitrates, NSAIDs, opioid analgesics, other antianginal drugs, other antidepressant drugs, selective serotonin reuptake inhibitors (SSRIs), skeletal muscle relaxants, and thiazide and related diuretics.

**Supplementary Note 2: Feature Attribution**

Feature attribution is a technique from explainable AI that attempts to explain how the inputs (clinical features) to a predictive model contribute to the output prediction.[62] There are a few approaches to defining feature attributions, but the most common approach is based on the concept of Shapley values from game theory.[15] Under the game-theoretic framework, a prediction, such as the risk of a negative outcome, is considered like a monetary debt. Each of the recorded features that were used to predict the patient's risk are then considered to be responsible for this debt, and as such, each feature is expected to contribute a number of points to the overall debt. Shapley values provide a mathematically formal way to compute how much each feature contributed to the debt, and therefore how much they contribute to the prediction.

Shapley values can be used even if their underlying predictive model is complex and nonlinear,[63] which has made them a popular tool for explaining the behavior of AI models.

The absolute risk prediction considers the predicted risk $F(X)$ given a vector of observed clinical features $X = (x_1, \ldots, x_M)$ for a certain patient. Clinical features, such as blood pressure, are not alone indicative of risk, but rather it is deviations from their typical values that may contribute to a risk prediction. Thus, we assume that for each clinical feature $x_i$, there is an associated reference value $\tilde{x}_i$ that encodes what we would expect to measure for an average healthy patient, represented by the vector $\tilde{X} = (\tilde{x}_1, \ldots, \tilde{x}_M)$. Baseline risk prediction considers how the risk $F(X)$ of an individual $X$ differs from the risk of the reference individual $F(\tilde{X})$.

Depending on what kind of predictive model is used, different algorithms might be preferred to calculate Shapley values. In this work, we compute Shapley values using the Kernel SHAP algorithm.[64] SHAP provides an efficient way to compute Shapley values for large models, since the baseline definition is too computationally intensive to compute directly. Kernel SHAP is particularly useful because it is widely applied, but also applicable to the tree-based models, such as Random Survival Forest, studied in our work.

When the function $F$ for predicting the risk is known (or assumed), then one can directly apply feature attribution methods to compute $\alpha_i$. For additive models, various feature attribution methods will yield the same result.[65] However, the generalization of $\alpha_i$ to non-additive models is not unique, and it requires a choice of definition. Generally, all such methods satisfy the completeness equation:

$$F(X) - F(\tilde{X}) = \sum_{i=1}^{M} \alpha_i \,. \quad (1)$$

Several popular approaches to feature attribution exist, and they differ in how they determine $\alpha_i$. The SHAP algorithm[64] is an extremely common choice, which extends the notion of Shapley values from binary to continuous inputs. SHAP defines the attribution coefficients $\alpha_i$ to be

$$\alpha_i = \sum_{\mathcal{S} \subseteq \mathcal{F} \setminus i} \frac{|\mathcal{S}|!\,(|\mathcal{F}| - |\mathcal{S}| - 1)!}{|\mathcal{F}|!} [F(\mathcal{S} \cup \{i\}) - F(\mathcal{S})], \quad (2)$$

where $\mathcal{F}$ is the set of all features, $F(\mathcal{S})$ is the model's predictions using only a subset $\mathcal{S} \subset \mathcal{F}$, and $|\mathcal{S}|$ is the number of elements in a set $\mathcal{S}$. When specific model architectures are chosen, there are specialized versions of the SHAP algorithm that allow SHAP values to be computed more efficiently.[64] Besides SHAP, there are several alternative methods for computing feature attributions, including LIME,[66] Integrated Gradients,[62] SmoothGrad,[17] and others.

There may be cases in which risk prediction is the goal, but there is no pre-established model to work with. In this case, the model should be inferred using approaches from machine learning and survival regression. Since many feature attribution methods are model agnostic, the choice of model is very flexible. In our experiments, we learn the unknown function $F$ using a Random Survival Forest[29] model. It offers robust predictions due to averaging over a number of survival trees, each fitted on different sub-samples of the data (sub-sample size matches the original dataset but the samples are drawn with replacement).